\documentclass[11pt,a4paper]{article}
\usepackage[hyperref,acceptedWithA]{tacl2018v2} %
\usepackage{times,latexsym}
\usepackage{url}
\usepackage[T1]{fontenc}
\usepackage{amsfonts}
\usepackage{amsmath}
\usepackage{microtype}
\usepackage{tipa}
\usepackage{booktabs}
\usepackage{multirow}
\usepackage{enumerate}

\usepackage[disable]{todonotes}
\makeatletter
\newcommand*\iftodonotes{\if@todonotes@disabled\expandafter\@secondoftwo\else\expandafter\@firstoftwo\fi}  %
\makeatother

\usepackage{cleveref}
\crefname{section}{\S}{\S\S}
\Crefname{section}{\S}{\S\S}
\crefname{table}{Tab.}{}
\crefname{figure}{Fig.}{}
\crefname{algorithm}{Algorithm}{}
\crefname{equation}{eq.}{}
\crefname{appendix}{App.}{}
\crefformat{section}{\S#2#1#3}  %

\newcommand{\plex}{p_{\textrm{\textit{lex}}}}
\newcommand{\qlex}{q_{\textrm{\textit{lex}}}}
\newcommand{\calQ}{\cal Q}

\newcommand{\xx}{\mathbf{x}}
\newcommand{\word}[1]{\textit{#1}}
\newcommand{\defn}[1]{\textbf{#1}}
\usepackage[small]{subcaption}
\usepackage{caption}
\title{Phonotactic Complexity and its Trade-offs}

\author{Tiago Pimentel\\University of Cambridge \\{\tt tp472@cam.ac.uk} \And Brian Roark\\Google\\{\tt roark@google.com}
\And Ryan Cotterell\\University of Cambridge\\{\tt rdc42@cam.ac.uk}
}

\date{}

\begin{document}
\maketitle
\begin{abstract}
We present methods for calculating a measure of phonotactic complexity---bits per phoneme---that permits
a straightforward cross-linguistic comparison. When
given a word, represented as a sequence of phonemic segments such as symbols
in the international phonetic alphabet, and a
statistical model trained on a sample of word types from the language, we
can approximately measure bits per phoneme using
the negative log-probability of that word
under the model. This simple measure allows
us to compare the entropy across languages, giving insight into how complex
a language's phonotactics are. Using a
collection of 1016 basic concept words across 106
languages, we demonstrate a very strong negative correlation of $-0.74$
between bits per phoneme and the average length of words.
\end{abstract}

\section{Introduction}
One prevailing view on system-wide phonological complexity is that as
one aspect increases in complexity (e.g., size of phonemic inventory),
another reduces in complexity (e.g., degree of phonotactic
interactions). Underlying this claim---the so-called compensation
hypothesis \cite{martinet,moran2014cross}---is the conjecture that
languages are, generally speaking, of roughly equivalent complexity,
i.e., no language is overall inherently more complex than another.
This conjecture is widely accepted in the literature and dates back,
at least, to the work of \newcite{hockett1958course}. Since
along any one axis, a language may be more complex than another,
this conjecture has a corollary that
compensatory relationships between
different types of complexity must exist. Such compensation has been hypothesized to be the result of
natural processes of historical  change, and is sometimes
attributed to a potential linguistic universal of equal communicative
capacity \cite{speech,coupe2019different}.

\begin{figure}
	\centering
        \includegraphics[width=\columnwidth]{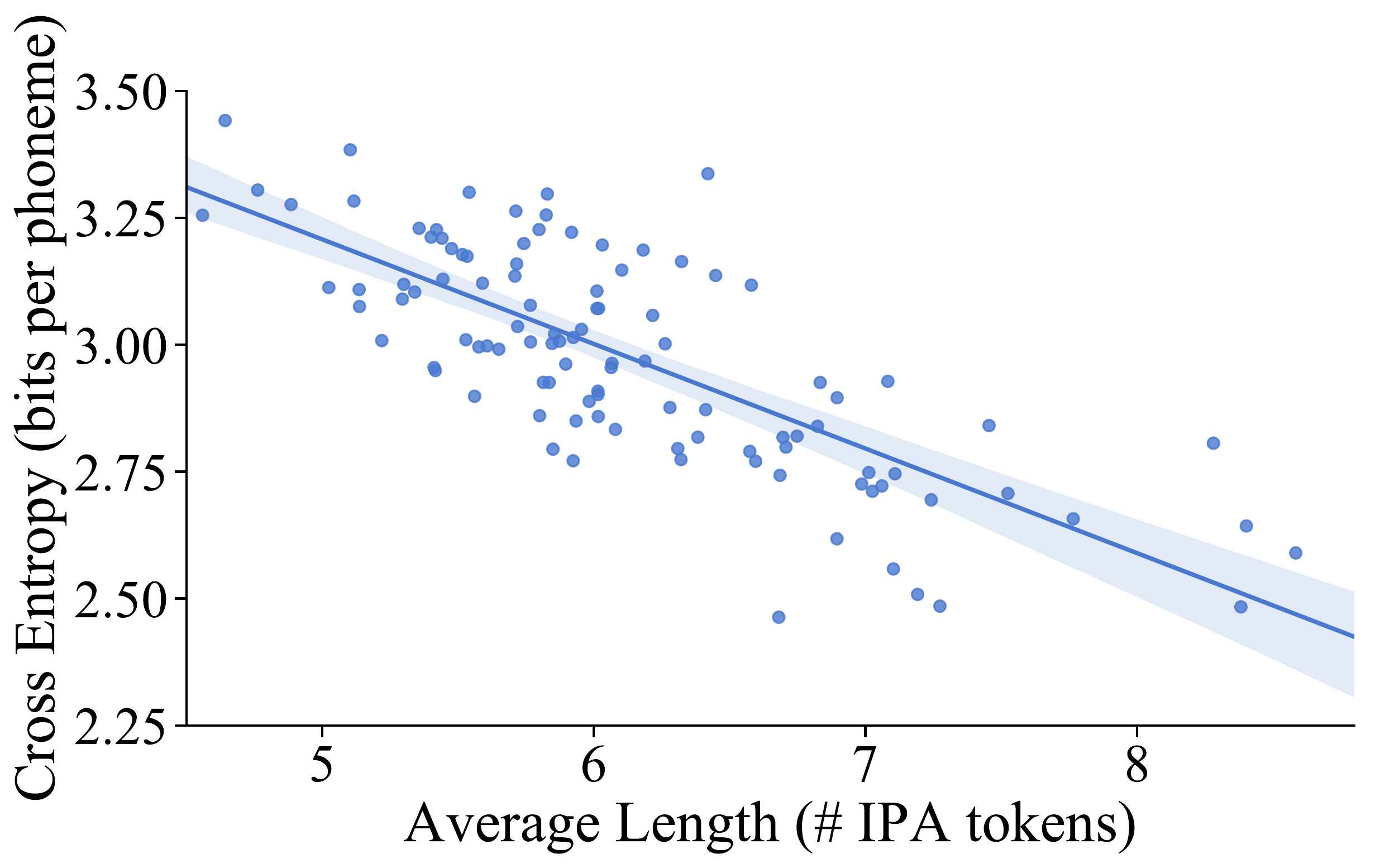}
        \caption{Bits-per-phoneme vs average word length using an LSTM language model.}
        \label{fig:init-full-corr}
\end{figure}

Methods for making hypotheses about linguistic complexity objectively
measurable and testable have long been of interest, though
existing measures are typically relatively coarse---see, e.g.,
\newcite{moran2014cross} and \cref{sec:background} below for reviews. 
Briefly, counting-based measures such as inventory sizes (e.g., numbers of vowels, consonants, syllables) typically play a key role in assessing phonological complexity.
Yet, in addition to their categorical nature, such measures generally do not capture longer-distance (e.g., cross-syllabic) phonological dependencies such as vowel harmony.
In this paper, we take an information-theoretic view of phonotactic
complexity, and advocate for a measure that permits straightforward
cross-linguistic comparison: bits per phoneme. For each language, a statistical language model over words (represented as phonemic sequences) is trained
on a sample of types from the language, and then used to calculate the bits per phoneme for new samples, thus providing an upper bound of the actual entropy \cite{brown1992estimate}. 

Characterizing phonemes via information theoretic measures goes back to \newcite{cherry1953toward}, who discussed the information content of phonemes in isolation, based on the presence or absence of distinctive features, as well as in groups, e.g., trigrams or possibly syllables.  Here we leverage modern recurrent neural language modeling methods to build models over full word forms represented as phoneme strings, thus capturing any dependencies over longer distances (e.g., harmony) in assigning probabilities to phonemes in sequence.  By training and evaluating on comparable corpora in each language, consisting of concept-aligned words, we can characterize and compare their phonotactics.  While probabilistic characterizations of phonotactics have been employed extensively in psycholinguistics (see \cref{sec:phonoprob}), such methods have generally been used to assess single words within a lexicon (e.g., classifying high versus low probability words during stimulus construction), rather than information-theoretic properties of the lexicon as a whole, which our work explores.\looseness=-1

The empirical portion of our paper exploits this information-theoretic take on complexity to examine multiple aspects of phonotactic complexity:
\begin{enumerate}[(i)]
\itemsep0em 
               \setlength{\leftmargin}{-.25in}
\item \textbf{Bits-per-phoneme and Word Length:}
In \cref{sec:hypothesis1}, we show a very high negative correlation of $-0.74$ between
bits-per-phoneme and average word length for the same 1016 basic concepts across 106 languages. 
This correlation is plotted in \cref{fig:init-full-corr}. In contrast, conventional phonotactic complexity measures, e.g., number of consonants in an inventory, demonstrate poor
correlation with word length. Our results are consistent with \newcite{speech}, who show a similar correlation in speech.\footnote{See also \newcite{coupe2019different}, where syllable-based bigram models are used to establish a comparable information rate in speech across 17 typologically diverse languages.} We additionally establish, in \cref{sec:control_confounds}, that the correlation persists when controlling for characteristics of long words, e.g., early versus late positions in the word.

\item \textbf{Constraining Language:}
Despite often being thought of as adding complexity, processes like vowel harmony and final-obstruent
devoicing \emph{improve} the predictability of subsequent segments by constraining the number of well formed forms.
Thus, they reduce complexity measured in bits per phoneme. 
We validate our models by systematically removing certain constraints in our corpora in \cref{sec:hypothesis3}.

\item \textbf{Intra- versus Inter-Family Correlation:}
Additionally, we present 
results in \cref{sec:hypothesis4} showing our complexity measure not only correlates with 
word length in a diverse set of languages, but also intra language 
families. Standard measures of phonotactic complexity do not 
show such correlations.
\item \textbf{Explicit feature representations:}
We also find (in \cref{sec:hypothesis5}) that methods for including features explicitly in the representation, using methods described in \cref{sec:models}, yield little benefit except in an extremely low-resource condition.\looseness=-1

\end{enumerate}

\noindent Our methods\footnote{Code to train these models and reproduce results is available at \url{https://github.com/tpimentelms/phonotactic-complexity}.} permit a straight-forward cross-linguistic comparison of phonotactic complexity, which we use to demonstrate an intriguing trade-off with word length.  Before motivating and presenting our methods, we next review related work on measuring complexity and phonotactic modeling. 

\section{Background: Phonological Complexity}\label{sec:background}
\subsection{Linguistic complexity}\label{sec:complexity}
Linguistic complexity is a nuanced topic. 
For example, one can judge a particular sentence to be syntactically complex relative to other sentences in the language. However, one can also describe a language as a whole as being complex in one aspect or another, e.g., polysynthetic languages are often deemed morphologically complex. In this paper, we look to characterize phonotactics at the language level. However, we use methods more typically applied to specific sentences in a language, for example in the service of psycholinguistic experiments.\looseness=-1

In cross-linguistic studies, the term {\it complexity\/} is generally used chiefly in two manners, which \newcite{moran2014cross} follow \newcite{miestamo2006feasability} in calling {\it relative\/} and {\it absolute\/}.  Relative complexity metrics are those that capture the difficulty of learning or processing language, which \newcite{miestamo2006feasability} points out may vary depending on the individual (hence, is relative to the individual being considered).  For example, vowel harmony, which we will touch upon later in the paper, may make vowels more predictable for a native speaker, hence less difficult to process; for a second language learner, however, vowel harmony may increase difficulty of learning and speaking.  Absolute complexity measures, in contrast, assess the number of parts of a linguistic (sub-)system, e.g., number of phonemes or licit syllables.

In the sentence processing literature, {\it surprisal} \cite{hale2001probabilistic,levy2008expectation} is a widely used measure of processing difficulty, defined as the negative log probability of a word given the preceding words. Words that are highly predictable from the preceding context have low surprisal, and those that are not predictable have high surprisal. 
The phonotactic measure we advocate for in \cref{sec:problex} is related to surprisal, though at the phoneme level rather than the word level, and over words rather than sentences. Measures related to phonotactic probability have been used in a range of psycholinguistic studies---see \cref{sec:phonoprob}---though generally to characterize single words within a language (e.g., high versus low probability words) rather than for cross-linguistic comparison as we are here.  Returning to the distinction made by \newcite{miestamo2006feasability}, we will remain agnostic in this paper as to which class (relative or absolute) such probabilistic complexity measures fall within, as well as whether the trade-offs that we document are bonafide instances of complexity compensation or are due to something else, e.g., related to the communicative capacity as hypothesized by \newcite{speech}.  We bring up this terminological distinction primarily to situate our use of \textit{complexity} within the diverse usage in the literature.

Additionally, however, we will point out that an important motivation for those advocating for the use of absolute over relative measures to characterize linguistic complexity in cross-linguistic studies is a practical one.
Miestamo \shortcite{miestamo2006feasability,miestamo2008grammatical} claims that relative complexity measures are infeasible for broadly cross-linguistic studies because they rely on psycholinguistic data, which is neither common enough nor sufficiently easily comparable across languages to support reliable comparison. In this study, we demonstrate that surprisal and related measures are not subject to the practical obstacles raised by Miestamo, independently of whichever class of complexity they fall into.

\subsection{Measures of Phonological Complexity}\label{sec:measures}
The complexity of phonemes has long been studied in linguistics,
including early work on the topic by \newcite{zipfpsycho}, who argued
that a phoneme's articulatory effort was related to its frequency.
\newcite{trubetzkoy} introduced the notion of markedness of
phonological features, which bears some indirect relation to both
frequency and articulatory complexity. Phonological complexity can be
formulated in terms of language production (e.g., complexity of
planning or articulation) or in terms of language processing (e.g.,
acoustic confusability or predictability), a distinction often framed
around the ideas of articulatory complexity and perceptual
salience---see, e.g., \newcite{maddieson2009calculating}. One recent instantiation
of this was the inclusion of both focalization and dispersion to
model vowel system typology \cite{cotterell-eisner-2017-probabilistic}.

It is also natural to ask questions about the phonological complexity of an entire language in addition to that of individual phonemes---whether articulatory or perceptual, phonemic or phonotactic.  Measures of such complexity that allow for
cross-linguistic comparison are non-trivial to define. We review several previously proposed metrics here. 

\paragraph{Size of phoneme inventory.}
The most basic metric proposed for measuring phonological complexity
is the number of distinct phonemes in the language's phonemic
inventory \cite{segmental}. There has been considerable historical
interest in counting both the number of vowels and the number of
consonants---see, e.g., \newcite{hockett1955manual,greenberg1978universals,maddieson1984patterns}. Phoneme inventory size has its limitations---it ignores
the phonotactics of the language. It does, however, have the advantage
that it is relatively easy to compute without further linguistic
analysis. Correlations between the size of vowel and consonant inventories
(measured in number of phonemes) have been extensively studied, with
contradictory results presented in the literature---see, e.g.,
\newcite{moran2014cross} for a review. Increases in phonemic inventory
size are also hypothesized to negatively correlate with word length
measured in phonemes \cite{moran2014cross}. In
\newcite{segmental}, an inverse relationship was demonstrated between
the size of the segmental inventory and the
mean word length for 10 languages, and similar results (with some qualifications) were
found for a much larger collection of languages in
\newcite{moran2014cross}.\footnote{Note that by examining negative
correlations between word length and inventory size within the context
of complexity compensation, word length is also being
taken implicitly as a complexity measure, as we shortly make explicit.}
We will explore phoneme inventory size as a baseline in
our studies in \cref{sec:experiments}.

\paragraph{Markedness in Phoneme Inventory.}
A refinement of phoneme inventory size takes
into account markedness of the individual phonemes. \newcite{mcwhorter2001world}
argues that one should judge the complexity of an inventory by counting
the cross-linguistic frequency of the phonemes in the inventory, channeling
the spirit of \newcite{greenberg1966language}. Thus, a language
that has fewer phonemes, but contains cross-linguistically marked ones
such as clicks, could be more complex.\footnote{\newcite{mcwhorter2001world} was one of the first to offer a quantitative treatment of linguistic
complexity at all levels. Note, however, he rejects the equal complexity
hypothesis, arguing creoles are simpler than other languages.  As our data contains no creole, we cannot address this hypothesis; rather we only compare non-creole languages.} \citeauthor{mcwhorter2001world} justifies this definition
with the observation that no attested language has a phonemic inventory
that consists only of marked segments. Beyond frequency, \newcite{lindblom1988phonetic}
propose a tripartite markedness rating scheme for various consonants.  In this paper, we are principally looking at phonotactic complexity, though we did 
examine the joint training of models across languages, which can be seen as modeling some degree of typicality and markedness.

\paragraph{Word length.}
As stated earlier, word length, measured in the number of phonemes in a word, has been shown to negatively correlate with other complexity measures, such as phoneme inventory size \cite{segmental, moran2014cross}.  To the extent that this is interpreted as being a compensatory relation, this would indicate that word length is being taken as an implicit measure of complexity.  Alternatively, word length has a natural interpretation in terms of information rate, so trade-offs could be attributed to communicative capacity \cite{speech,coupe2019different}.

\paragraph{Number of Licit Syllables.}
Phonological constraints extend beyond individual units to the
structure of entire words themselves, as we discussed above; so why
stop at counting phonemes? One step in that direction is to
investigate the syllabic structure of language, and count the number
of possible licit syllables in the language
\cite{maddieson1984patterns,shosted2006correlating}. Syllabic
complexity brings us closer to a more holistic measure of phonological
complexity. Take, for instance, the case of Mandarin Chinese. At first
blush, one may assume that Mandarin has a complex phonology due to an
above-average-sized phonemic inventory (including tones);
closer inspection, however, reveals a more constrained system.
Mandarin only admits two codas:
/\textipa{n}/ and /\textipa{N}/.

While syllable inventories and syllable-based
measures of phonotactic complexity---e.g., highest complexity
syllable type in \newcite{maddieson2006}---
do incorporate more of the constraints at play in a language versus segment-based measures, (a) they remain relatively simple counting measures; and (b) phonological constraints do not end at the syllable boundary. Phenomena such as vowel
harmony operate at the word level. Further, the combinatorial possibilities captured by a syllabic inventory, as discussed by \newcite{maddieson2009calculating}, can be seen as a sort of categorical version of a distribution over forms. Stochastic models of
word-level phonotactics permit us to go beyond simple enumeration of a
set, and characterize the distribution in more robust
information-theoretic terms.

\subsection{Phonotactics}
Beyond characterizing
the complexity of phonemes in isolation or the number of syllables,  one can also look at
the system determining how phonemes combine to form longer sequences
in order to create words. The study of which sequences of phonemes constitute natural-sounding
words is called phonotactics. For example, as \newcite{chomskyhalle1965}
point out in their oft-cited example, {\it brick\/} is an actual
word in English;\footnote{For
  convenience, we just use standard orthography to represent
  actual and possible words, rather than phoneme strings.} {\it blick\/} is not
an actual word in English, but is judged to be a possible word
by English speakers; and {\it bnick\/} is neither an actual nor a
possible word in English, due to constraints on its phonotactics.\looseness=-1

Psycholinguistic studies often use phonotactic probability to characterize stimuli within a language---see \cref{sec:phonoprob} for details.  For example, \newcite{goldrick2008phonotactic}
demonstrate that both articulatory complexity and phonotactic probability influence the speed and accuracy of speech production. Measures of the overall complexity of a phonological system must thus also account for phonotactics.\looseness=-1

\newcite{cherry1953toward} took an explicitly information-theoretic view of phonemic structure, including discussions of both encoding phonemes as feature bundles and the redundancy within groups of phonemes in sequence. This perspective of phonemic coding has led to work on characterizing the explicit rules or constraints that lead to redundancy in phoneme sequences, including morpheme structure rules \cite{halle59} or conditions \cite{stanley1967redundancy}. Recently, \newcite{futrell2017generative} take such approaches as inspiration for a generative model over feature dependency graphs.  We, too, examine decomposition of phonemes into features for representation in our model (see \cref{sec:models}), though in general this only provided modeling improvements over atomic phoneme symbols in a low-resource scenario.\looseness=-1

Much of the work in phonotactic modeling is intended to explain the sorts of grammaticality judgments exemplified by the examples of \newcite{chomskyhalle1965} discussed earlier. Recent work is typically founded on the commonly held perspective that such judgements are gradient\footnote{Gradient judgments would account for the fact that 
  {\it bwick\/} is typically judged to be a possible English word like {\it
    blick\/} but not as good. In other words, {\it bwick\/} is better than {\it
    bnick\/} but not as good as {\it blick\/}.} hence amenable to
stochastic modeling, e.g., \newcite{hayes2008maximum} and
\newcite{futrell2017generative}---though cf.
\newcite{gorman2013generative}. In this paper, however, we are looking
at phonotactic modeling as the means for assessing phonotactic
complexity and discovering potential evidence of trade-offs cross-linguistically, and are not strictly speaking evaluating the model on its ability to capture such judgments, gradient or otherwise.\looseness=-1

\subsection{Phonotactic probability and surprisal}\label{sec:phonoprob}
A word's phonotactic probability has been shown to influence both
processing and learning of language. Words with high phonotactic
probabilities (see brief notes on the operationalization of this below) have
been shown to speed speech processing, both recognition
\cite[e.g.,][]{vitevitch1999probabilistic} and production
\cite[e.g.,][]{goldrick2008phonotactic}. Phonotactically probable
words in a language have also been shown to be easier to learn
\cite[{\it inter
  alia}]{storkel2001learning,storkel2003learning,coady2004young},
although such an effect is also influenced by {\it
  neighborhood density\/} \cite{coady2003phonological}, as are the
speech processing effects \cite{vitevitch1999probabilistic}.
Informally, phonological neighborhood density is the number of similar
sounding words in the lexicon, which, to the extent that high
phonotactic probability implies phonotactic patterns frequent in the
lexicon, typically correlates to some degree with phonotactic
probability---i.e., dense neighborhoods will typically consist of
phonotactically probable words. Some effort has been made to
disentangle the effect of these two characteristics
\cite[{\it inter alia}]{vitevitch1999probabilistic,storkel2006differentiating,storkel2011independent}.

Within the psycholinguistics literature referenced above, phonotactic
probability was typically operationalized by summing or averaging
the frequency with which single phonemes and phoneme bigrams occur, either
overall or in certain word positions (initial, medial, final); and
neighborhood density of a word is typically the number of words in a lexicon
that have Levenshtein distance 1 from the word \cite[see,
e.g.,][]{storkel2010online}. Note that these measures are used to
characterize specific words, i.e., given a lexicon, these measures
allow for the designation of high versus low phonotactic probability
words and high versus low neighborhood density words, which is useful for designing experimental stimuli. Our
bits-per-phoneme measure, in contrast, is used to characterize the
distribution over a sample of a language rather than specific individual words in
that language.\looseness=-1

Other work has made use of phonotactic probability to examine how such
processing and learning considerations may impact the lexicon.
\newcite{dautriche2017words} take phonotactic probability as one
component of ease of processing and learning---the other being
perceptual confusability---that might influence how lexicons become
organized over time. They operationalize phonotactic probability via
generative phonotactic models (phoneme $n$-gram models and
probabilistic context-free grammars with syllable structure), hence
closer to the approaches described in this paper than the work cited
earlier in this section. Generating artificial lexicons from such
models, they find that real lexicons demonstrate higher network
density (as indicated by Levenshtein distances, frequency of minimal
pairs, and other measures) than the randomly generated lexicons,
suggesting that the pressure towards highly clustered lexicons is
driven by more than just phonotactic probability.

Evidence of pressure towards communication efficiency in the lexicon
has focused on both phonotactic probability and word length. The
information content, as measured by the probability of a word in
context, is shown to correlate with orthographic length (taken as a
proxy for phonological word length)
\cite{piantadosi2009communicative,piantadosi2011word}.
\newcite{piantadosi2012communicative} show that words with lower bits
per phoneme have higher rates of homophony and polysemy, in support of
their hypothesis that words that are easier to process will have
higher levels of ambiguity. Relatedly, \newcite{mahowald2018word}
demonstrate, in nearly all of the 96 languages investigated, a high
correlation between orthographic probability (as proxy for phonotactic
probability) and frequency, i.e., frequently used forms tend to be
phonotactically highly probable, at least within the word lengths
examined (3-7 symbols). A similar perspective on the role of
predictability in phonology holds that words that are high probability
in context (i.e., low surprisal) tend to be reduced, and those that
are low probabilty in context are prone to change
\cite{hume2013role} or to some kind of enhancement
\cite{hall2018role}. As \newcite{priva2018interdependence} point out,
frequency, predictabilty and information content (what they call {\it
  informtivity\/} and operationalize as expected predictability) are
related and easily confounded, hence the perspectives presented by
these papers are closely related. Again, for these studies and those
cited earlier, such measures are used to characterize individual words
within a language rather than the lexicon as a whole.

\section{The Probabilistic Lexicon}\label{sec:problex}
In this work, we are interested in a hypothetical phonotactic
distribution $\plex : \Sigma^* \rightarrow \mathbb{R}_+$ over the
lexicon. In the context of phonology, we interpret $\Sigma^*$ as all
``universally possible phonological surface forms,'' following
\newcite{hayes2008maximum}.\footnote{\newcite{hayes2008maximum} label
  $\Sigma^*$ as $\Omega$.} The distribution $\plex$, then, assigns a
probability to every possible surface form $\xx \in \Sigma^*$.  In the
special case that $\plex$ is a log-linear model, then we arrive at
what is known as a maximum entropy grammar
\cite{goldwater2003learning,jager2007maximum}. A good distribution
$\plex$ should assign high probability to phonotactically valid words,
including non-existent ones, but little probability to phonotactic
impossibilities. For instance, the possible English word \word{blick}
should receive much higher probability than
\word{$^*$bnick}, which is not a possible English word.
The lexicon of a language, then, is considered to be
generated as samples without replacement from $\plex$.\looseness=-1

If we accept the existence of the distribution $\plex$, then a
natural manner by which we should measure the phonological complexity of language is
through Shannon's entropy \cite{cover2012elements}:
\begin{equation}\label{eq:entropy}
H(\plex) = -\sum_{\xx \in \Sigma^*} \plex(\xx) \log \plex(\xx)
\end{equation}
The units of $H(\plex)$ are bits as we take $\log$ to be base 2. Specifically,
we will be interested in \defn{bits per phoneme}, that is,
how much information each phoneme in a word conveys, on average. 

\subsection{Linguistic Rationale}
Here we seek to make a linguistic argument for the adoption of bits
per phoneme as a metric for complexity in the phonological
literature. Bits are fundamentally units of predictability: If the
entropy of your distribution is higher, that is more bits, then it is
\emph{less} predictable, and if the entropy is lower, that is, fewer
bits, then it is \emph{more} predictable with an entropy of 0
indicating determinism.

\paragraph{Holistic Treatment.}
When we just count the number of distinctions in individual parts of
the phonology, e.g., number of vowels or number of consonants, we do
not get a holistic picture of how these pieces interact. A simple
probabilistic treatment will \emph{inherently} capture nuanced
interactions. Indeed, it is not clear how to balance
the number of consonants, the number of vowels and the number
of tones to get a single number of phonological complexity.
Probabilistically modeling phonological strings, however,
does capture this. We judge the complexity of a phonological
system as its entropy.\looseness=-1

\paragraph{Longer-Distance Dependencies.}
To the best of the authors' knowledge, the largest phonological unit
that has been considered in the context of cross-linguistic phonological complexity is the syllable,
as discussed in \cref{sec:measures}. However, the syllable clearly
has limitations. It cannot capture, tautologically, cross-syllabic
phonological processes, which abound in the languages of the world.
For instance, vowel and consonant harmony are quite common
cross-linguistically. Naturally, a desideratum for any
measure of phonological complexity is to consider all levels of phonological
processes. Examples of vowel
harmony in Turkish are presented in \cref{tab:harmony}.\looseness=-1

\begin{table}
\centering
\small
  \begin{tabular}{l|lll|l} \toprule
  \textbf{English} & \textbf{Turkish} && \textbf{English} & \textbf{Turkish} \\
  \midrule
    ear & kulak & \hspace*{0.1in} &
    throat & bo\u{g}az \\
    rain & ya\u{g}mur &  &
    foam & k{\"o}p{\"u}k \\
    summit & zirve &  &
    claw & pen\c{c}e \\
    nail & t\i rnak &  &
    herd & s{\"u}r{\"u} \\
    horse & beygir &  &
    dog & k{\"o}pek \\ \bottomrule
  \end{tabular}
  \caption{Turkish evinces two types of vowel harmony, front-back and round-unround. Here we focus on just front-back harmony. The examples in the table above are such that all vowels in a word are either back (\i, u, a, o) or front (i, {\"u}, e, {\"o}), which is generally the case.}
  \label{tab:harmony} 
\end{table}
 
\paragraph{Frequency Information.}
None of the previously proposed phonological complexity measures deals with the
fact that certain patterns are more frequent than others; probability
models inherently handle this as well. Indeed, consider the role of the
unvoiced velar fricative /x/ in English; while not part of the
canonical consonant inventory, /x/ nevertheless appears in a variety
of loanwords. For instance, many native English speakers do pronounce
the last name of composer Johann Sebastian Bach as /bax/.
Moreover, English phonology acts upon /x/ as one would
expect: consider Morris Halle's \shortcite{halle1978knowledge} example \word{Sandra out-Bached
  Bach}, where the second word is pronounced /out-baxt/ with a final
/t/ rather than a /d/.  We conclude that /x/ is in the
consonant inventory of at least some native English speakers. However,
counting it on equal status with the far more common /k/ when
determining complexity seems incorrect. Our probabilistic metric
covers this corner case elegantly. 

\paragraph{Relatively Modest Annotation Requirements.}
Many of these metrics require a linguist's analysis of the
language. This is a tall order for many languages. Our probabilistic
approach only requires relatively simple annotations, namely, a
\newcite{swadesh1955towards}-style list in the international phonetic alphabet (IPA) to estimate a distribution. When
discussing why he limits himself to counting complexities, \newcite{maddieson2009calculating} writes:

\begin{quotation}
\noindent ``[t]he factors considered in these studies
only involved the inventories of consonant and vowel contrasts, the
tonal system, if any, and the elaboration of the syllable canon. It is
relatively easy to find answers for a good many languages to such
questions as `how many consonants does this language distinguish?' or
`how many types of syllable structures does this language allow?' ''
\end{quotation}
The moment one searches for data on more elaborate notions of
complexity, e.g., the existence of vowel harmony, one is faced
with the paucity of data---a linguist must have analyzed
the data.

\subsection{Constraints Reduce Entropy}
Many phonologies in the world employ hard constraints, e.g., a syllable final obstruent must be devoiced or the vowels in a word must be harmonic. Using our definition of phonological complexity as entropy, we can 
prove a general result that \emph{any} hard-constraining process will reduce entropy, thus, making the phonology \emph{less} complex. The fact that this holds for any hard contraint, be it vowel harmony or final-obstruent devoicing, is a fact that conditioning reduces entropy. 

\subsection{A Variational Upper Bound}
If we want to compute \cref{eq:entropy}, we are immediately
faced with two problems. First, we do not know $\plex$: we simply
assume the existence of such a distribution from which the words
of the lexicon were drawn. Second, even if we did know $\plex$,
computation of the $H(\plex)$ would be woefully intractable, as it
involves an infinite sum. Following \newcite{brown1992estimate},
we tackle both of these issues together. Note that this line of reasoning
follows \newcite{cotterell-etal-2018-languages} and \newcite{mielke-etal-2019-kind} who use a similar technique for measuring language complexity
at the sentence level.

We start with a basic inequality from information theory.
For any distribution $\qlex$ with the same support as $\plex$,
the cross-entropy provides an upper bound on the entropy,
i.e.,
\begin{equation}\label{eq:inequality}
H(\plex) \leq H(\plex, \qlex)
\end{equation}
where cross-entropy is defined as
\begin{equation}\label{eq:cross-entropy}
  H(\plex, \qlex) = -\sum_{\xx \in \Sigma^*} \plex(\xx) \log \qlex(\xx)
\end{equation}
Note that \cref{eq:inequality} is tight if and only if $\plex = \qlex$.
We still are not done, as \cref{eq:cross-entropy} still requires
knowledge of $\plex$ and involves an infinite sum. However,
we are now in a position to exploit samples from $\plex$. Specifically,
given $\tilde{\xx}^{(i)} \sim \plex$, we approximate
\begin{equation}\label{eq:empirical}
  H(\plex, \qlex) \approx -\frac{1}{N} \sum_{i=1}^N \log \qlex(\tilde{\xx}^{(i)})
\end{equation}
with equality if we let $N \rightarrow \infty$. In information theory, this equality
in the limit is called the asymptotic equipartition property
and follows easily from the weak law of large numbers.
Now, we have an empirical procedure for estimating an upper bound on $H(\plex)$.
For the rest of the paper, we will use the right-hand side of
\cref{eq:empirical} as a surrogate for the phonotactic complexity
of a language. 

\paragraph{How to choose $\qlex$?}
Choosing a good $\qlex$ is a two-step process. First, we choose
a variational family $\calQ$. Then, we choose a specific $\qlex \in \calQ$
by minimizing the right-hand side of \cref{eq:empirical}
\begin{equation}
 \qlex = \text{argsup}_{q \in \calQ} \frac{1}{N} \sum_{i=1}^N \log q(\tilde{\xx}^{(i)})
\end{equation}
This procedure corresponds to maximum likelihood estimation. 
In this work, we consider two variational families: (i)
a phoneme $n$-gram model, and (ii) a phoneme-level RNN
language model. We describe each in \cref{sec:models}.

\subsection{A Note on Types and Tokens}
To make the implicit explicit, in this work we will exclusively
be modeling types, rather than tokens. We briefly justify this discussion from both theoretical and practical concerns. From a theoretical side, a token-based model is unlikely to correctly model an OOV distribution as very frequent tokens often display unusual phonotactics for historical reasons. A classic example comes from English: consider the appearance of /\textipa{D}/. Judging by token-frequency, /\textipa{D}/ is quite common as it starts some of the most common words in the language: \word{the}, \word{they}, \word{that}, etc.  However, novel words categorically avoid initial 
/\textipa{D}/. From a statistical point of view,
one manner to justify type-level modeling is through the Pitman--Yor process \cite{ishwaran2003generalized}. \newcite{goldwater2006interpolating} showed that type-level modeling is a special case of the stochastic process, writing that they ``justif[y] the appearance of type frequencies in formal analyses of natural language.''

Practically, using token-level frequencies, even in a dampened form, is not possible due to the large selection of languages we model. Most of the languages we consider do not have corpora large enough to get reasonable token estimates. Moreover, as many of the languages we consider have a small number of native speakers, and, in extreme cases, are endangered, the situation is unlikely to remedy itself, forcing the phonotactician to rely on types.\looseness=-1

\section{Methods}
\subsection{Phoneme-Level Language Models}\label{sec:models}

\paragraph{Notation.}
Let $\Sigma$ be a discrete alphabet of symbols from the international phonetic alphabet (IPA), including special beginning-of-string and end-of-string symbols. A character level language model (LM) models a probability distribution over $\Sigma^*$:
\begin{equation}
  p(\xx) = \prod\limits_{i=1}^{|\xx|} p \left(x_i \mid \xx_{<i} \right)
\end{equation}

\paragraph{Trigram LM.}
$n$-grams assume the sequence follows a $(n\!-\!1)$-order Markov model, conditioning the probability of a phoneme on the $(n\!-\!1)$ previous ones
\begin{equation}
f_{\mathit{n}}(x_i \mid \xx_{<i}) = \frac{\mathit{count}(x_i, x_{i-1},
  \dots, x_{i+1-n})}{\mathit{count}(x_{i-1}, \dots, x_{i+1-n})}
\end{equation}
where we assume the string $\mathbf{x}$ is properly padded with beginning and end-of-string symbols. 

The trigram model used in this work is estimated as the deleted interpolation \cite{jelinek1980interpolated}  of the trigram, bigram and unigram relative frequency estimates\looseness=-1
\begin{equation}
p_{\textit{3}} (x_i \mid \xx_{<i}) = \sum_{n=1}^3 \alpha_n f_{n}(x_i \mid \xx_{<i})
\end{equation}
where the mixture parameters $\alpha_n$ were estimated via Bayesian optimization with a Gaussian prior maximizing the expected improvement on a
validation set, as discussed by \newcite{snoek2012practical}.

\paragraph{Recurrent Neural LM.}
Recurrent neural networks excel in language modeling, being able to
capture complex distributions $p(x_i \mid \xx_{<i})$ \cite{mikolov2010recurrent,sundermeyer2012lstm}. Empirically, recent work
has observed dependencies on up
to around 200 tokens \cite{khandelwal2018sharp}. We use a
character-level Long Short-Term Memory \citep[LSTM,][]{hochreiter1997long} language model, which is the state of the art for
character-level language modeling \cite{merity2018analysis}.

Our architecture receives a sequence of tokens $\xx \in \Sigma^*$ and embeds each token $x_i \in
\Sigma$ using a dictionary-lookup embedding
table. This results in vectors $z_i \in \mathbb{R}^{d}$ which are fed
into an LSTM. This LSTM produces a high-dimensional representation of
the sequence, often termed hidden states:\looseness=-1
\begin{equation}
  h_i = \textit{LSTM}\left(z_{i-1}, h_{i-1}\right) \in \mathbb{R}^{d}
\end{equation}
These representations are then fed into a softmax to produce a 
distribution over the next character:
\begin{equation}
  p\left(x_{i} \mid  \xx_{< i}\right) = \textit{softmax}\left(W h_i + b\right)
\end{equation}
where $W \in \mathbb{R}^{|V| \times d}$ is a final projection matrix and $b \in \mathbb{R}^{|\Sigma|}$ is a 
bias term. In our implementation, $h_0$ is a vector of all zeros and $z_0$ is the lookup embedding for the beginning-of-string token.

\paragraph{Phoneme Embedding LM.}
When developing a phoneme-level recurrent neural LM, one can use a base of phonemic features---e.g. Phoible \citep{phoible}---to implement a multi-hot embedding such that similar phonemes will have similar embedding representations. A phoneme $i$ will have a set of binary attributes $a_{i}^{(k)}$ (e.g. stress, sonorant, nasal), each with its corresponding embedding representation $z^{(k)}$. A phoneme embedding will, then, be composed by the element-wise average of each of its features lookup embedding
\begin{equation}
    z_i = \frac{\sum_{k} a_i^{(k)} z^{(k)}}{\sum_{k} a_i^{(k)}}
\end{equation}
 where $a_i^{(j)}$ is $1$ if phoneme $i$ presents attribute $j$ and $z^{(j)}$ is the lookup embeddings of attribute $j$.
This architecture forces similar phonemes, measured in terms of overlap in distinctive features, to have similar representations. %

\subsection{NorthEuraLex Data}\label{sec:data}
We make use of data from the NorthEuraLex corpus \cite{northeuralex}.
The corpus is a concept-aligned multi-lingual lexicon with data
from 107 languages. The lexicons contains 1016 ``basic'' concepts.
Importantly, NorthEuraLex is appealing for our study as all the words are written
in a unified IPA scheme. A sample of the lexicon is provided in \cref{tab:lexicon}.
For the results reported in this paper, we omitted Mandarin, since no tone information was included in its annotations, causing its phonotactics to be greatly underspecified.  No other tonal languages were included in the corpus, so all reported results are over 106 languages.

\paragraph{Why is Base-Concept Aligned Important?}
Making use of data that is concept aligned across the languages provides a certain amount of control (to the extent possible) of the influence of linguistic content on the forms that we are modeling.  In other words, these forms should be largely comparable across the languages in terms of how common they are in the active vocabulary of adult speakers.  Further, base concepts as defined for the collection are more likely to be lemmas without inflection, thus reducing the influence of morphological processes on the results.\footnote{Most of the concepts in the dataset do not contain function words and verbs are in the bare infinitive form -- e.g., {\it have}, instead of {\it to have}) -- although there are a few exceptions. For example, the German word {\it hundert} is represented as {\it a hundred} in English.}

To test this latter assertion, we made use of the UniMorph\footnote{https://unimorph.github.io}
morphological database \cite{kirov-etal-2018-unimorph} to look up words and assess the percentage that correspond to lemmas or base forms. Of the 106 languages in our collection, 48 are also in the UniMorph database, and 46 annotate their lemmas in a way that allowed for simple string matching with our word forms.  For these 46 languages,  on average we found 313 words in UniMorph of the 1016 concepts (median 328). A mean of 87.2\% (median 93.3\%; minimum 58.6\%) of these matched lemmas for that language in the UniMorph database.  This rough string matching approach provides some indication that the items in the corpus are largely composed of such base forms.

\begin{table}\small
\resizebox{\columnwidth}{!}{%
  \begin{tabular}{llll} \toprule
  \textbf{Concept} & \textbf{Language} & \textbf{Word} & \textbf{IPA} \\
  \midrule
    eye & portuguese & olho & /\textipa{oLu}/ \\
	ear & finnish & korva & /\textipa{kOrVA}/ \\
	give & north karelian & antua & /\textipa{AntUA}/ \\
	tooth & veps & hambaz & /\textipa{hAmbAz}/ \\
	black & northern sami & \u{c}\'{a}hppes & /\textipa{\t{\textteshlig}aahppes}/ \\
	immediately & hill mari & t\"{o}p\"{o}k & /\textipa{t\o{}r\o{}k}/ \\ \bottomrule
  \end{tabular}}
  \caption{Sample of the lexicon in NorthEuraLex corpus.}
  \label{tab:lexicon}
\end{table}

\paragraph{Dataset Limitations.}
Unfortunately, there is less typological diversity in our dataset than
we would ordinarily desire. NorthEuraLex draws its languages from
21 distinct language families that are spoken in Europe and
Asia. This excludes languages indigenous to the
Americas,\footnote{Inuit languages, which are genetically related to
  the languages of Siberia, are included in the lexicon.} Australia,
Africa and South-East Asia. While lamentable, we know of no other concept-aligned lexicon 
 with broader typological diversity that is written in a unified phonetic alphabet, so we must save studies of more typologically diverse set of languages for future work. 

In addition, we note that the process of base concept selection and identification of corresponding forms from each language \cite[detailed in][]{dellert15,dellert2017information} was non-trivial, and some of the corpus design decisions may have resulted in somewhat biased samples in some languages.  For example, there was an attempt to minimize the frequency of loanwords in the dataset, which may make the lexicons in loanword heavy languages, such as English with its extensive Latinate vocabulary, somewhat less representative of everyday use than in other languages.  Similarly, the creation of a common IPA representation over this number of languages required choices that could potentially result in corpus artifacts.  As with the issue of linguistic diversity, we acknowledge that the resource has some limitations but claim that it is the best currently available dataset for this work.

\begin{figure*}
	\centering
        \includegraphics[width=\textwidth]{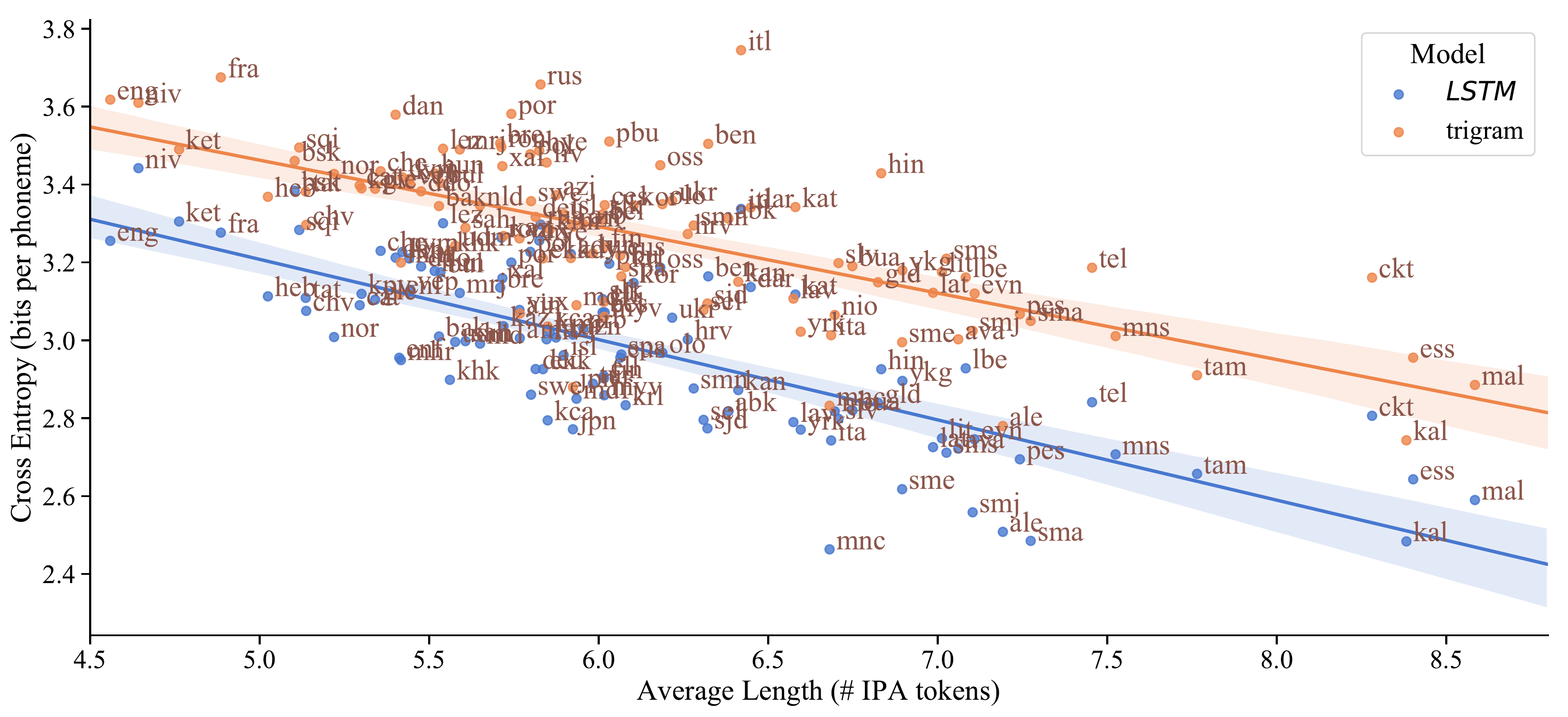}
        \caption{Per-phoneme complexity vs average word length under both a trigram and an LSTM language model.} \label{fig:lm_complexity}\label{fig:full-corr}
\end{figure*}

\begin{figure}
	\centering
        \includegraphics[width=\columnwidth]{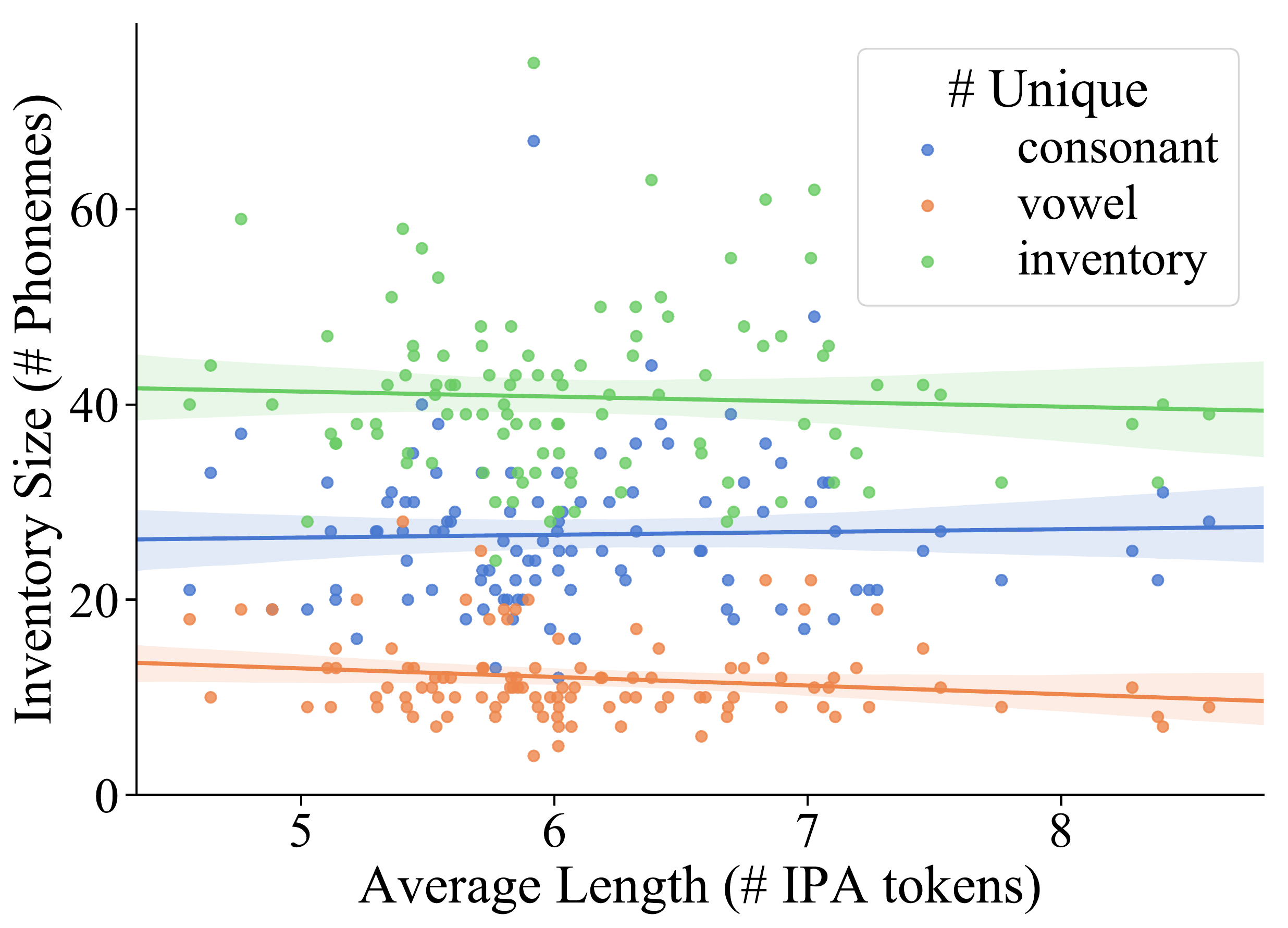}
        \caption{Conventional measures of phonological complexity vs average word length. These complexity measures are based in inventory size.} \label{fig:inventory_complexity}\label{fig:inventory-corr}
\end{figure}

\paragraph{Splitting the Data.}
We split the data at the concept level into 10 folds, used for cross validation. We create train-dev-test splits where the training portion has 8 folds ($\approx 812$ concepts) and the dev and test portions have 1 fold each ($\approx 102$ concepts).
We then create language-specific sets with the language-specific words for the concept to be rendered. Cross-validation allows us to have all 1016 concepts in our test sets (although evaluated using different model instances), and we do our following studies using all of them.

\subsection{Artificial Languages}\label{sec:artificial}
In addition to naturally occurring languages, we are also interested in artificial ones.
Why? We wish to validate our models in a controlled setting, quantifying the contribution of specific linguistic phenomena to our complexity measure. Thus,
developing artificial languages, which only differ with respect to one phonological property is useful.

\paragraph{The Role of Final-Obstruent Devoicing.}
Final-obstruent devoicing
\emph{reduces} phonological complexity under our information-theoretic
metric. The reason is simple: there are fewer valid syllables as all
those with voiced final obstruents are ruled out. Indeed, this point
is also true of the syllable counting metric discussed in
\cref{sec:measures}.
One computational notion of complexity
might say that the complexity of the phonology is equal to the number of states required to encode the transduction
from an underlying form to a surface form in a minimal finite-state transduction. 
Note that all SPE-style rules may be so encoded \cite{kaplan1994regular}. Thus,
the complexity of the phonotactics could be said to be related to the number of SPE-style rules that operate.
In contrast, under our metric, any process that constrains the number of possibilities will, inherently,
reduce complexity.  The studies in \cref{sec:hypothesis3} allow us to examine the \emph{magnitude} of such a reduction, and validate our models with respect to this expected behavior.

We create two artificial datasets without final-obstruent devoicing
based on the German and Dutch portions of NorthEurLex. We reverse the
process, using the orthography as a guide. For example, the German
/\texttslig u\textlengthmark k/ is converted to /\texttslig
u\textlengthmark g/ based on the orthography \word{Zug}.

\paragraph{The Role of Vowel Harmony.}
Like final obstruent devoicing, vowel harmony plays a roll in reducing the number of licit syllables.
In contrast to final obstruent devoicing, however, vowel harmony acts cross-syllabically.
Consider the Turkish lexicon, where most, but not all basic lexical items obey vowel harmony.
Processes like this reduce the entropy of $\plex$ and, thus, can be considered
as creating a less complex phonotactics.

For vowel harmony, we create 10 artificial datasets by randomly
replacing each vowel in a word with a new sampled (with replacement)
vowel from that language's vowel inventory. This breaks all vowel
harmony, but keeps the syllabic structure.

\begin{table}
\centering
\begin{small}
\begin{tabular}{l@{~~~~~~}cc}
 \toprule
& \multicolumn{2}{c}{\textbf{Correlation}}\\ \cmidrule{2-3}
\textbf{Measure} & Pearson $r$ & Spearman $\rho$\\
\midrule
Number of: \\
~~~~~~~~~~phonemes &  -0.047 & -0.054 \\
~~~~~~~~~~vowels &  -0.164 & -0.162 \\
~~~~~~~~~~consonants &  0.030 & 0.045 \\
\hline
Bits/phoneme: \\
~~~~~~~~~~unigram & -0.217 & -0.222 \\
~~~~~~~~~~trigram & -0.682 & -0.672 \\
~~~~~~~~~~LSTM & -0.762  & -0.744 \\
 \bottomrule
\end{tabular}
\end{small}
\caption{Pearson and Spearman rank correlation coefficients between complexity measures
and average word length in phoneme segments.}\label{tab:correlation}
\end{table}

\section{Results}\label{sec:experiments}

\subsection{Study 1: Bits Per Phoneme Negatively Correlates with Word Length}\label{sec:hypothesis1}
As stated earlier, \newcite{speech} investigated a complexity trade-off with
the information density of speech. From a 7-language study
they found a strong correlation
($R=-0.94$) between the information density and the syllabic complexity
of a language. One hypothesis adduced to explain these findings
is that, for functional reasons, the rate of linguistic information is
very similar cross-linguistically. Inspired by their study, we
conduct a similar study with our phonotactic setup. We hypothesize that the bits per phoneme for a given concept correlates with the number of phonemes in the word. Moreover, the bits per word should be similar across languages. 

\begin{figure}[t]
	\centering
        \includegraphics[width=\columnwidth]{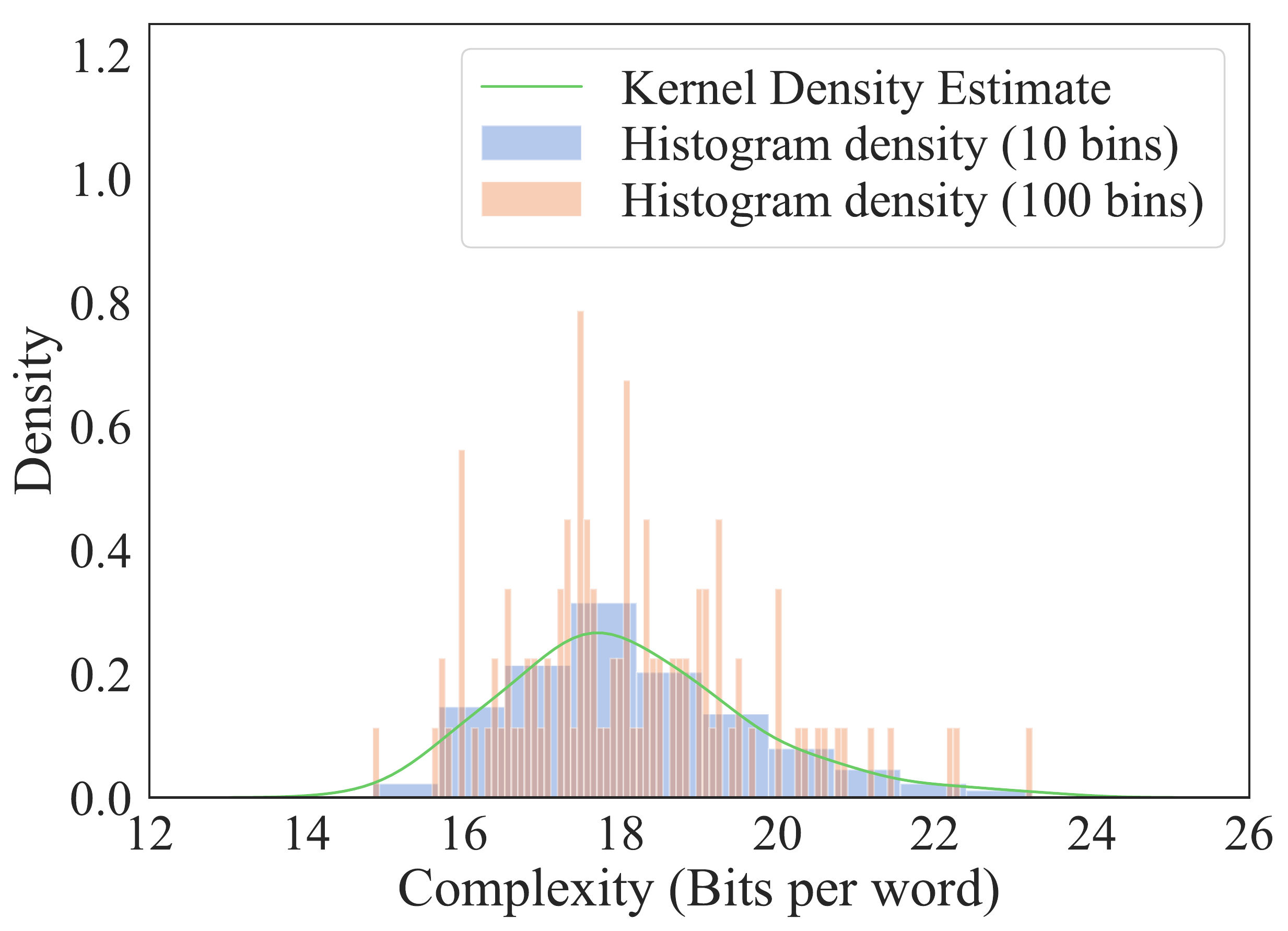}
        \caption{Kernel density estimate (KDE) of the average phonotactic complexity per word across 106 different languages. Different languages tend to present similar complexities (bits per word).} \label{fig:kde-word-complexity}
\end{figure}

We consider the relation between the average bits
per phoneme of a held-out portion of a language's lexicon, as measured
by our best language model, and the average length of the words in that language.
We present the results in Figures \ref{fig:lm_complexity} and \ref{fig:inventory_complexity} and in \cref{tab:correlation}.
We find a strong correlation under the LSTM LM (Spearman's $\rho=-0.744\textit{ with }p < 10^{-19}$).
At the same time, we only see a weak correlation under conventional measures of phonotactic complexity, such as vowel inventory size (Spearman's $\rho=-0.162\textit{ with }p = 0.098$). In \cref{fig:kde-word-complexity}, we plot the kernel density estimate and histogram densities (both 10 and 100 bins) of word-level complexity (bits per word).

\subsection{Study 2: Possible Confounds for Negative Correlations}\label{sec:control_confounds}

One possible confound for these results is that phonemes later in a word may in general have higher probability given the previous phonemes than those earlier in the string. This sort of positional effect was demonstrated in Dutch \cite{son2003information}, where  position in the word accounted for much of the variance in segmental information.\footnote{We briefly note that the \newcite{son2003information} study did not make use of a train/dev/test split of their data, but rather simply analyzed raw relative frequency over their Dutch corpus. As a result, all positions beyond any word onset that is unique in their corpus would have probability one, leading to a more extreme position effect than we would observe using regularization and validating on unseen forms.}  To ensure that we are not simply replicating such a positional effect across many languages, we performed several additional analyses.

\paragraph{Truncated words.} First, we calculated the bits-per-phoneme for just the first three positions in the word, and then looked at the correlation between this word-onset bits-per-phoneme and the average (full) word length in phoneme segments.  In other words, for the purpose of calculating bits-per-phoneme, we truncated all words to a maximum of three phonemes, and in such a way explicitly eliminated the contribution of positions later in any word.  Using the LSTM model, this yielded a Spearman correlation of $\rho=-0.469$ ($p < 10^{-7}$) , in contrast to $\rho=-0.744$ without such truncation (reported in \cref{tab:correlation}).  This suggests that there is a contribution of later positions to the effect presented in \cref{tab:correlation} that we lose by eliding them, but that even in the earlier positions of the word we are seeing a trade-off with full average word length.

\paragraph{Correlation with phoneme position.} We next looked to measure a position effect directly, by calculating the correlation between word position and bits for that position across all languages.  Here we find a Spearman correlation of $\rho=-0.429$ ($p<10^{-200}$), which again supports the contention that later positions in general require fewer bits to encode. Nonetheless, this correlation is still weaker than the per-language word length one (of $\rho=-0.744$).

\paragraph{Per-word correlations.} We also calculated the correlation between word length and bits per phoneme across all languages (without averaging per language here). The Spearman correlation between these factors---at the word level using all languages---is $\rho=-0.312$ ($p < 10^{-19}$). Analyzing each language individually, there is an average Spearman's $\rho=-0.257$ ($p < 10^{-19}$) between bits per phoneme and word length.  The minimum negative (i.e., highest magnitude) correlation of any language in the set is $\rho=-0.607$.  These per word correlations are reported in the upper half of \cref{tab:correlation_lang_effect}.

\begin{table}[t]
\centering
\begin{small}
\begin{tabular}{l@{~~~}cc}
 \toprule
& \multicolumn{2}{c}{\textbf{Correlation}}\\ \cmidrule{2-3}
\textbf{Measure} & Pearson $r$ & Spearman $\rho$\\
\midrule
Per Word: \\
~~~~~~all languages & -0.269 & -0.312 \\
~~~~~~each language (avg) &  -0.220 & -0.257 \\
~~~~~~each language (min) &  -0.561 & -0.607 \\
\hline
Per Language: \\
~~~~~~Fake (avg) & -0.270 & -0.254 \\
~~~~~~Fake (min) & -0.586 & -0.568 \\
~~~~~~Real & -0.762  & -0.744 \\
 \bottomrule
\end{tabular}
\end{small}
\caption{Pearson and Spearman rank correlation coefficients between complexity measures and word length in phoneme segments. All correlations are statistically significant with $p<10^{-8}$.}\label{tab:correlation_lang_effect}
\end{table}

\paragraph{Permuted `language' correlations.} Finally, to determine if our language effects perhaps arise due to the averaging of word lengths and bits per phoneme for each language, we ran a permutation test on languages. We shuffle words (with their pre-calculated bits-per-phoneme values) into 106 sets with the same size as the original languages---thus creating fake `languages'. We take the average word length and bits per phoneme in each of these fake languages and compare the correlation---returning to the `language' level this time---with the original correlation. After running this test for $10^4$ permutations, we found no shuffled set with an equal or higher Spearman (or Pearson) correlation than the real set. Thus, with a strong confidence ($p < 10^{-4}$) we can state there is a language level effect.  Average and minimum negative correlations for these `fake' languages (as well as the real set for ease of comparison) are presented in the lower half of \cref{tab:correlation_lang_effect}.

\begin{table}
\centering
\begin{small}
\begin{tabular}{@{}lccc} \toprule
& \multicolumn{2}{c}{\textbf{Complexity}}\\
\cmidrule{2-3}
\textbf{Model} & Orig & Art & \textbf{Diff} \\
\midrule
trigram: \\
~~~~~~~~~~German & 3.703 & 3.708 & 0.005 (0.13\%)~ \\
~~~~~~~~~~Dutch & 3.607 & 3.629 & 0.022 (0.58\%)$^\dagger$ \\
\hline
LSTM: \\
~~~~~~~~~~German & 3.230 & 3.268 & 0.038 (1.18\%)$^\dagger$ \\
~~~~~~~~~~Dutch & 3.161 & 3.191 & 0.030 (0.95\%)$^\dagger$ \\ \bottomrule
\end{tabular}
\end{small}
\caption{Complexities for original and artificial languages when removing final-obstruent devoicing. $^\dagger$~represents an statistically significant difference with $p<0.05$}\label{tab:artificial}
\end{table}

\subsection{Study 3: Constraining Languages Reduces Phonotactic  Complexity}\label{sec:hypothesis3}
Final-obstruent devoicing and vowel harmony reduce the number of licit syllables in a language, hence reducing the entropy. 
To determine the magnitude that such effects can have on the measure for our different model types,
we conduct two studies. In the first, we remove final-obstruent devoicing from the German and Dutch languages
in NorthEuraLex, as discussed in \cref{sec:artificial}. In the second study, we remove vowel harmony from 10 languages that have it,\footnote{The languages with vowel harmony are: bua, ckt, evn, fin, hun, khk, mhr, mnc, myv, tel, and tur.} as also explained in \cref{sec:artificial}.

\begin{figure}
\centering
	\includegraphics[width=2.6in]{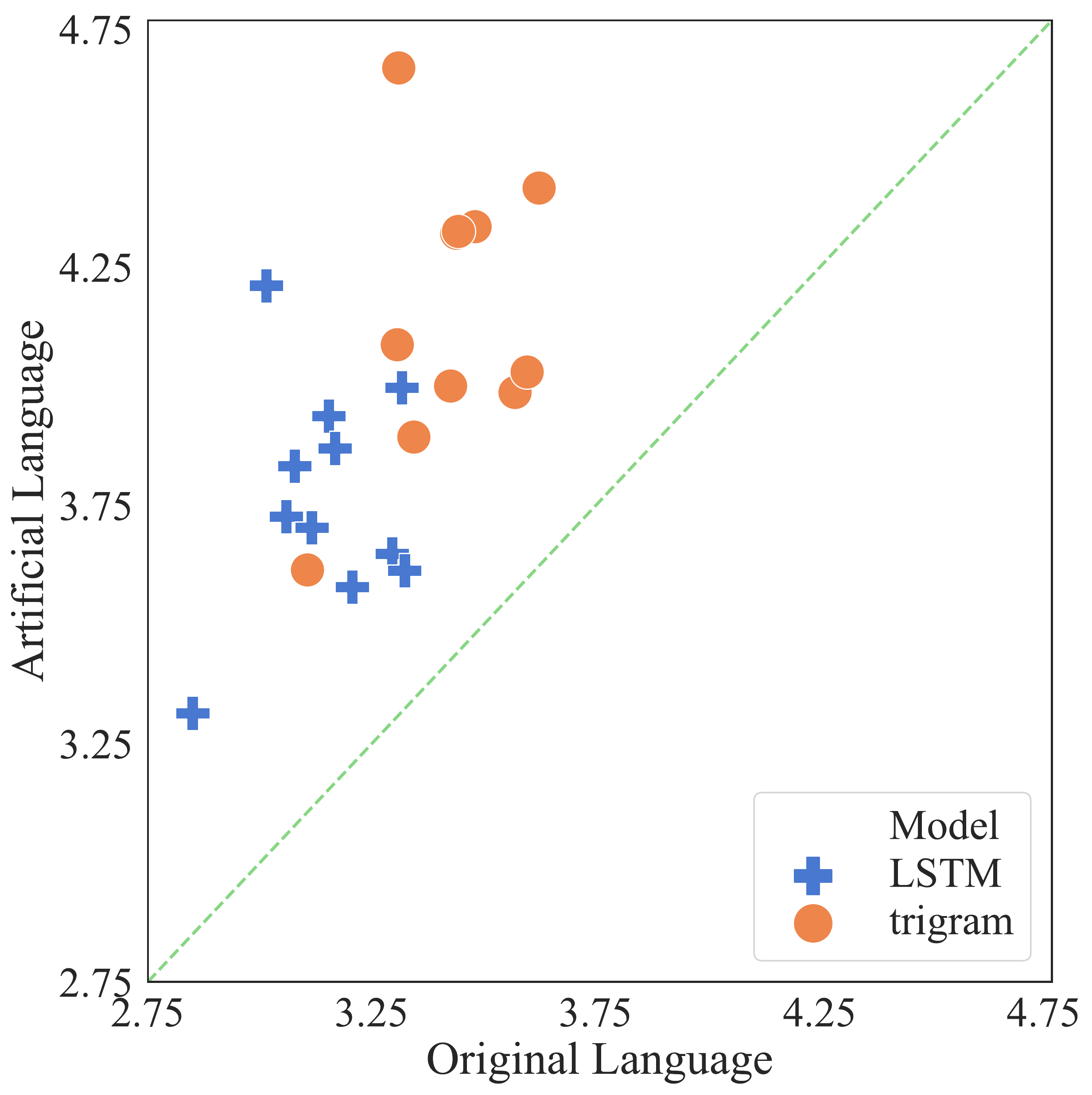}
\caption{Complexities for natural and artifical languages when removing vowel harmony. A paired permutation test showed all differences present statistical difference with $p < 0.01$.}\label{tab:artificial-harmony}
\end{figure}

After deriving two artificial languages without obstruent devoicing from both German and Dutch, we used 10 fold cross validation to train models for each language. The statistical relevance of differences between normal and artificial languages was analyzed using paired permutation tests between the pairs. Results are presented in \cref{tab:artificial}. We see that the $n$-gram can capture this change in complexity for Dutch, but not for German.
At the same time, the LSTM shows a statistically significant increase of $\approx0.034$ bits per phoneme when we remove obstruent devoicing from both languages.  Fig. \ref{tab:artificial-harmony} presents a similar impact on complexity from vowel harmony removal, as evidenced by the fact that all points fall above the equality line. Average complexity increased by $\approx 0.62$ bits per phoneme (an approximate $16\%$ entropy increase), as measured by our LSTM models.

In both of these artificial language scenarios, the LSTM models appeared more sensitive to the constraint removal, as expected.

\subsection{Study 4: Negative Trade-off Persists Within and Across Families}\label{sec:hypothesis4}
\newcite{moran2014cross} investigated the correlation between the number of phonological units in a language and its average word length across a large and varied set of languages. They found that, while these measures of phonotactic complexity (number of vowels, consonants or phonemes in a language) are correlated with word length when measured \textit{across} a varied set of languages, such a correlation usually does not hold \textit{within} language families. We hypothesize that this is due to their measures being rather coarse approximations to phonotactic complexity, so that only large changes in the language would show significant correlation given the noise. We also hypothesize that our complexity measure is less noisy, hence should be able to yield significant correlations both within and across families.

Results in \cref{tab:correlation} show a strong correlation for the LSTM measure, while they show a weak one for conventional measures of complexity. As stated before, \newcite{moran2014cross} found that vowel inventory size shows a strong correlation to word length on a diverse set of languages, but, as mentioned in  \cref{sec:data}, our dataset is more limited than desired. To test if we can mitigate this effect we average the complexity measures and word length per family (instead of per language) and calculate the same correlations again. These results are presented in \cref{tab:correlations_lstm_family_avg} and show that when we average these complexity measures per family we indeed find a stronger correlation between vowel inventory size and average word length, although with a higher null hypothesis probability (Spearman's $\rho=-0.367\textit{ with }p = 0.111$). We also see our LSTM based measure still shows a strong correlation (Spearman's $\rho=-0.526\textit{ with }p = 0.017$).

\begin{table}
\centering
\begin{small}
\begin{tabular}{@{}lcc} \toprule
& \multicolumn{2}{c}{\textbf{Correlation}}\\
\cmidrule{2-3}
\textbf{Measure} & Pearson $r$ & Spearman $\rho$\\
\midrule
Number of: \\
~~~~~~~~~~phonemes &  -0.214 & -0.095 \\
~~~~~~~~~~vowels &  -0.383 & -0.367 \\
~~~~~~~~~~consonants &  -0.147 & -0.092 \\
\hline
Bits/phoneme: \\
~~~~~~~~~~unigram & -0.267 & -0.232 \\
~~~~~~~~~~trigram & -0.621 & -0.520 \\
~~~~~~~~~~LSTM & -0.778  & -0.526 \\ \bottomrule
\end{tabular}
\end{small}
\caption{Pearson and Spearman correlation between complexity measures
and word length in phoneme segments averaged across language families.}\label{tab:correlations_lstm_family_avg}
\end{table}

We now analyze these correlations intra families, for all family languages in our dataset with at least 4 languages. These results are presented in \cref{tab:intra_family_p}.
Our LSTM based phonotactic complexity measure shows strong intra family correlation with average word length for all five analyzed language families ($-0.662 \geq \rho \geq -1.0\textit{ with }p < 0.1$).
At the same time, vowel inventory size only shows a negative statistically significant correlation within Turkic.

\begin{table}
\centering
\begin{small}
\begin{tabular}{lccc} \toprule
& \multicolumn{2}{c}{\textbf{Spearman $\rho$}} & \\
\cmidrule{2-3}
\textbf{Family} & \textbf{LSTM} & \textbf{Vowels} & \textbf{\# Langs} \\
\midrule
Dravidian & -1.0$^*$ & -0.894~~ & 4 \\
Indo-European & -0.662$^*$ & -0.218~~ & 37 \\
Nakh-Daghestanian & -0.771$^{\dagger}$ & -0.530~~ & 6 \\
Turkic & -0.690$^{\dagger}$ & -0.773$^{\dagger}$ & 8 \\
Uralic & -0.874$^*$ &  ~0.363$^{\dagger}$ & 26 \\
\multicolumn{4}{@{}p{2in}}{\footnotesize $^*$ Statistically significant with $p<0.01$} \\
\multicolumn{4}{@{}p{2in}}{\footnotesize $^\dagger$ Statistically significant with $p<0.1$} \\ \bottomrule
\end{tabular}
\end{small}
\caption{Spearman correlation between complexity measures
and average word length per language family. Phonotactic complexity in bits per phoneme presents very strong intra-family correlation with word length in three of the five families. Size of vowel inventory presents intra-family correlation in Turkic and Uralic.}\label{tab:intra_family_p}
\end{table}

\subsection{Study 5: Explicit feature representations do not generally improve models}\label{sec:hypothesis5}

\cref{tab:correlation} presents strong correlations when using an LSTM with standard one-hot lookup embedding.  Here we train LSTMs with three different phoneme embedding models: (1) a typical Lookup embedding, in which each Phoneme has an associated embedding; (2) a phoneme features based embedding, as explained in \cref{sec:models}; (3) the concatenation of the Lookup and the Phoneme embedding. We also train these models both using independent models for each language, and with independent models, but sharing embedding weights across languages.

We first analyze these model variants under the same lens as used in Study 1. \cref{tab:complexities_vs_corr} shows the correlations between the complexity measure resulting from each of this models and the average number of phonemes in a word. We find strong correlations for all of them ($-0.740 \geq \rho \geq -0.752\text{ with }p < 10^{-18}$). 
We also present in \cref{tab:complexities_vs_corr} these models' cross entropy, averaged across all languages.  At least for the methods that we are using here, we derived no benefit from either more explicit featural representations of the phonemes or by sharing the embeddings across languages.

\begin{table}[t]
\centering
\begin{small}
\begin{tabular}{@{}lcc@{}} \toprule
\textbf{Model} & \textbf{Complexity} & \textbf{Spearman $\rho$} \\
\midrule
$n$-Grams: \\
~~~~~~~~~~unigram  & 4.477 & -0.222 \\
~~~~~~~~~~trigram  & 3.270 & -0.672 \\
\hline
Independent Embeddings: \\
~~~~~~~~~~Lookup & 2.976 & -0.744 \\
~~~~~~~~~~Phoneme & 2.992 & -0.741 \\
~~~~~~~~~~Lookup + Phoneme & 2.975 & -0.752 \\
\hline
Shared Embeddings: \\
~~~~~~~~~~Lookup & 2.988 & -0.743 \\
~~~~~~~~~~Phoneme & 2.977 & -0.744 \\
~~~~~~~~~~Lookup + Phoneme & 2.982 & -0.740 \\ \bottomrule
\end{tabular}
\end{small}
\caption{Average cross-entropy across all languages and the correlation between complexity and average word length for different models.} \label{tab:complexities_vs_corr}
\end{table}

We also investigated scenarios using less training data, and it was only in very sparse scenarios (e.g., using just 10\% of the training used in our standard trials, or 81 example words) where we observed even a small benefit to explicit feature representations and shared embeddings.

\section{Conclusion}
We have presented methods for calculating a well-motivated measure of phonotactic complexity: bits per phoneme. This measure is derived from information theory and its value is calculated using the probability distribution of a language model.  We demonstrate that cross-linguistic comparison is straightforward using such a measure, and find a strong negative correlation with average word length.  This trade-off with word length can be seen as an example of complexity compensation or perhaps related to communicative capacity.

\section*{Acknowledgments}
We thank Dami\'an E. Blasi for his feedback on previous versions of this paper and the anonymous reviewers, as well as action editor Eric Fosler-Lussier, for their constructive and detailed comments---the paper is much improved as a result.

\bibliography{phono_complexity}
\bibliographystyle{acl_natbib}

\end{document}